\documentclass[11pt]{article}
\newif\ifarxiv
\arxivtrue

\usepackage{acl}

\usepackage{xurl}
\usepackage{times}
\usepackage{latexsym}
\usepackage[T1]{fontenc}
\usepackage[utf8]{inputenc}
\usepackage{inconsolata}
\usepackage{graphicx}
\graphicspath{{figures/}}
\usepackage{placeins}
\usepackage{booktabs}
\usepackage{tabularx}
\usepackage{ragged2e}
\usepackage{array}
\usepackage{threeparttable}
\usepackage{enumitem}
\usepackage{amsmath}
\usepackage{amssymb}
\usepackage{longtable}
\DeclareUnicodeCharacter{2013}{--}
\DeclareUnicodeCharacter{2014}{--}
\DeclareUnicodeCharacter{2019}{'}
\DeclareUnicodeCharacter{2018}{`}
\DeclareUnicodeCharacter{201C}{``}
\DeclareUnicodeCharacter{201D}{''}
\DeclareUnicodeCharacter{2026}{\ldots}
\DeclareUnicodeCharacter{00A0}{~}
\DeclareUnicodeCharacter{2265}{\ensuremath{\geq}}
\DeclareUnicodeCharacter{2248}{\ensuremath{\approx}}
\DeclareUnicodeCharacter{2208}{\ensuremath{\in}}
\DeclareUnicodeCharacter{2212}{\ensuremath{-}}
\DeclareUnicodeCharacter{2032}{\ensuremath{^\prime}}
\DeclareUnicodeCharacter{03BA}{\ensuremath{\kappa}}
\DeclareUnicodeCharacter{0394}{\ensuremath{\Delta}}
\DeclareUnicodeCharacter{03C6}{\ensuremath{\phi}}
\DeclareUnicodeCharacter{00D7}{\ensuremath{\times}}
\DeclareUnicodeCharacter{2713}{\checkmark}
\DeclareUnicodeCharacter{2717}{x}
\makeatletter
\AtBeginDocument{\setlength{\@fptop}{0pt}\setlength{\@fpbot}{0pt plus 1fil}}
\makeatother
\AtBeginDocument{%
  \setlength{\abovedisplayskip}{4pt}%
  \setlength{\belowdisplayskip}{4pt}%
  \setlength{\abovedisplayshortskip}{2pt}%
  \setlength{\belowdisplayshortskip}{2pt}%
  \setlength{\jot}{3pt}%
}

\title{FrIdéo: French News Outlet Ideology Scoring via Multi-Source Evidence Fusion and Distancing-Aware Lexical Scoring}

\author{Amr Sobhy \\
  Le French News Lab \\
  \texttt{amr@frenchnewslab.org}}

\begin{document}
\maketitle

\begin{abstract}

Outlet ideology is a standard explanatory variable in media framing, media effects, and political communication research, but continuous, ecosystem-wide coverage lets it enter as graded exposure rather than a coarse bin. For France, existing references are either broad but non-analytic or analytic but sparse and source-specific, and none we are aware of offers ecosystem-wide coverage on a unified continuous scale. We contribute FrIdéo: to our knowledge the first ecosystem-wide continuous ideology scores for the French press, a transparent composite score for 30 French news outlets, fusing nine independent families of evidence with empirical-Bayes-style shrinkage toward a structural prior. For the text-based family we introduce \emph{symmetric plain/qualified scoring}, which separates the plain use of ideologically loaded vocabulary from its qualified (distanced) use on both sides, a distinction bag-of-words methods miss. The resulting scale runs from L'Humanité on the far left to Valeurs actuelles on the far right; the far-right pole is near-indistinguishable and CNews ties Le Figaro. Because the text family reads current articles, it also captures editorial change that static sources miss, such as the \emph{Journal du Dimanche}'s sharp rightward shift after its 2023 ownership change. The ordering is robust to perturbation and to dropping any single family, and agrees with held-out Media Bias Fact Check labels. We release the scores, confidence intervals, and replication code.

\end{abstract}

\section{Introduction}

The political orientation of news outlets is among the most consequential variables in political communication research. It shapes event selection, framing, source amplification, and audience belief updating. Continuous, ecosystem-wide scores let ideology enter as a graded exposure rather than a coarse bin \citep{martinYurukoglu2017}, and avoid selecting on whichever outlets a rating panel happens to cover. Yet scoring outlet ideology is methodologically fraught: outlets resist self-categorization, audience surveys conflate perceived vs.\ actual bias, and content analyses require large corpora and fine-grained annotation.

The dominant paradigms in the English-language literature fall into four camps. \emph{Audience-based methods} infer outlet ideology from the partisanship of its readers or social-media followers \citep{bakshy2015}. \emph{Content-based methods} analyze word choice, topic emphasis, or citation patterns, or crowd-code article content, to place outlets on a liberal--conservative axis \citep{grosecloseMilyo2005,gentzkowShapiro2010,budak2016}. \emph{Network-based methods} derive positions from citation, follower, or web-link graphs linking outlets to political actors \citep{cointet2021,fondationDescartes2020}. \emph{External-rating methods} rely on crowd-sourced or expert panels such as Media Bias Fact Check (MBFC) or AllSides \citep{mbfc2026}. Each has known failure modes: audience methods conflate readership ideology with editorial stance; content methods require large corpora and struggle with outlets that report without endorsing; network methods inherit the snapshot date of the graph they are built on; rating panels are U.S.-centric and sparsely cover European media.

For the French media ecosystem, the situation is acute. France has dozens of nationally relevant news outlets spanning print, broadcast, and digital-native formats, covering the full spectrum from the Communist \emph{L'Humanité} to the identitarian \emph{Valeurs actuelles}. Available references for outlet ideology in France are either broad but non-analytic (descriptive typologies without a measurement model) or analytic but sparse and source-specific: \citet{pew2018} place only eight outlets on a perceived-ideology scale; \citet{cage2024} estimate partisan deviation for broadcast outlets only, from TV and radio guest bookings; \citet{cointet2021} derive continuous positions via network analysis but from a static 2018--2019 snapshot; \citet{bensonHallin2007} offer a comparative content analysis of only three outlets. No single source currently combines broad ecosystem coverage with a unified continuous scale for the sampled French national news ecosystem.

This paper contributes a resource: FrIdéo, to our knowledge the first continuous, ecosystem-wide ideology dataset for French national news, a latent, anchor-free, z-standardized score with a per-outlet 95\% bootstrap confidence interval (CI) for 30 outlets, together with the method used to build it. It is validated through cross-family convergence, known-groups separation, and robustness checks; held-out MBFC agreement is secondary external corroboration, not the target being reconstructed. Two methodological components make it possible: a unified multi-source framework that fuses nine evidence families into a single latent score via per-family z-standardization and empirical-Bayes shrinkage toward a structural prior; and a signal-based lexical score (D7) that runs on raw article text and distinguishes \emph{plain use} of ideologically loaded vocabulary (no explicit distancing signal present) from \emph{qualified use} (guillemets, distancing-prefix, or side-specific critique vocabulary present), yielding a time-sensitive editorial signal that registers within-period editorial shifts (e.g., JDD's 2023 realignment) unrecoverable from static structural priors or historical typologies; validated against a manually adjudicated gold standard it reaches an operational-fidelity $\kappa$ = 0.938 (0.78 on human-adjudicated rows alone), exceeding both an LLM judge ($\kappa$ = 0.49) and a CamemBERT probe (macro-F1 = 0.775) on this narrow signal-detection task. We follow the outlet-scoring literature in treating ideology as a single left--right continuum. This is an explicit modeling choice, which we test rather than assume (Section~\ref{sec:results}, Appendix~\ref{appendix:unidimensional}), and which can collapse distinct economic and cultural axes where these diverge. We conjecture the distancing check transfers to other languages with quotation conventions and a lexicalized political register, but D7 is validated on French items only, so that transfer is untested. Unlike phrase-co-occurrence and citation-pattern approaches that anchor to a reference corpus of ideologically tagged speech \citep{gentzkowShapiro2010,grosecloseMilyo2005}, FrIdéo imposes no external anchor. FrIdéo is a resource-building step, not a definitive settlement.

\section{Related Work}

\citet{grosecloseMilyo2005} and \citet{gentzkowShapiro2010} estimate U.S. outlet slant from citation patterns and phrase co-occurrence with congressional speech, but both require large ideological reference corpora that do not transfer to French media. \citet{budak2016} crowd-code article content; \citet{ronnback2025} train a classifier on outlet-level GDELT features with MBFC ground-truth labels. FrIdéo differs by treating outlet ideology as a \emph{composite latent position} derived from nine heterogeneous evidence families rather than optimizing a single content model, making each source's directional contribution transparent and accommodating sparse content coverage, at the cost of multi-source evidence assembly. Event-level framing work~\citep{zhao2025,das2024,conchaMacias2024} addresses the distinct problem of how a given event is framed across outlets, complementing our outlet-level scores.

\paragraph{Editorial change over time.} Most outlet-scoring work treats ideology as static within the estimation window, but a smaller literature does not: \citet{martinYurukoglu2017} recover MSNBC's post-2006 leftward move from time-varying slant estimates, and \citet{kim2022dynamic} show slant moves substantially even over short horizons. That outlets shift is therefore established; our contribution is integrating it into a multi-source model, pairing one within-period text family with eight static ones so an ownership change registers in the fused score instead of being averaged away. For French we are aware of no existing time-varying measure.

For the French context, \citet{bensonHallin2007}, working within the comparative-media-systems tradition, characterize the French press as more closely tied to the political field than its US counterpart. \citet{cage2022,cage2024} provide among the most rigorous empirical evidence for French outlet ideology, using panel data on TV and radio guest bookings to estimate partisan deviation relative to a public-broadcaster benchmark. \citet{cointet2021} apply correspondence analysis to the Twitter-follower network of 831 MPs to place outlets on an economic left-right axis. \citet{pew2018} provides survey-based perceived-ideology scores, and \citet{fondationDescartes2020} maps the French politico-media space from Twitter follower-interaction networks. \citet{rebillardSklower2022} and \citet{cage2017} document ownership concentration and its political implications. Our framework synthesizes all of these into a common scale.

Lexical content-scoring methods often confuse \emph{reporting about} partisan vocabulary with \emph{plain use} of it \citep[for a review of bias-detection methods, see][]{hamborg2019}, since bias is often sentence-level \citep{lim2020}: an outlet that covers far-right demonstrations must mention \emph{ensauvagement} (`descent into savagery') or \emph{grand remplacement} (`great replacement') to describe them; counting these as plain use inflates its right score. Our plain/qualified procedure (Section~5) applies a three-layer distancing check before crediting an occurrence. We adapt James--Stein shrinkage \citep{jamesStein1961,efronMorris1977}, in the spirit of latent ideal-point estimation in political science \citep{jackman2001,shorMcCarty2011}, to heterogeneous evidence with variable coverage: rich-evidence outlets weight toward the empirical mean, sparse ones toward a historical-tradition prior.

\section{The French Outlet Panel}

Our panel consists of 30 French-language national news outlets: 15 national newspapers and magazines, 6 broadcast TV/radio news channels, 8 digital-native outlets, and 1 cooperative regional paper (see Table~\ref{tab:fullmodel} for the full list). The selection aimed for broad coverage of the national news ecosystem rather than a threshold-filtered sample. Inclusion criteria were: national or near-national reach; a general political or current-affairs mandate (excluding sports, lifestyle, and trade outlets); representation of all major formats (daily and weekly press, broadcast, digital-native, regional press); and coverage of the full ideological spectrum from far-left (\emph{L'Humanité}) to identitarian far-right (\emph{Fdesouche}, \emph{Valeurs actuelles}). No minimum audience threshold was applied. No stratum is an exhaustive enumeration of its format: coverage of the ACPM-ranked national dailies is complete (below), while the magazine, broadcast (6) and digital-native (8) strata are purposive selections. \emph{Ouest-France} is the single regional title because it outsells every national paper in the same ACPM ranking; its cooperative ownership is incidental, not the inclusion criterion. The panel includes every title in the ACPM (Alliance pour les Chiffres de la Presse et des Médias) 2025 paid-circulation and 2026 readership rankings\footnote{ACPM \emph{Diffusion Presse Quotidienne Nationale} 2025 and \emph{OneNext PQN} 2026-S1, \url{https://www.acpm.fr/classements}.} except the sports daily \emph{L'Équipe} and the international edition of the \emph{New York Times}, both outside our general-news, French scope; \emph{L'Opinion} is additionally included as a national daily below ACPM's reporting threshold.

The panel includes outlets under a range of ownership structures: state/public, cooperative/non-profit, legacy independent, and oligarchic (Bolloré/Vivendi, Arnault/LVMH, Dassault, Drahi, Pinault). Ownership is treated as one evidence family (D0) rather than as a model constraint; removing D0 entirely yields leave-one-out (LOO) $\rho = 0.997$, confirming it is not load-bearing.

\section{The Ideology Scoring Framework}
\label{sec:framework}

\subsection{Evidence Families}

We define nine direct indicator families (with D6 split into two sub-families, D6a and D6b) and one structural prior (S2). Table~\ref{tab:families} summarizes each family.

Source consistency parameters $\sigma_k$ are estimated via leave-one-family-out residual variance: $\sigma_k^2 = \mathrm{Var}(z_{ik} - \bar{z}_{i \setminus k})$ across outlets with data in k, where $\bar{z}_{i \setminus k}$ is the mean of all other families. Lower $\sigma_k$ means the family tracks the aggregate closely; higher $\sigma_k$ reflects divergence---noise or genuine signal absent from the consensus. This is a modeling assumption, not a ground-truth criterion. Families capturing signal absent from historical sources---notably D7, the only within-period lexical family---exhibit larger $\sigma_k$ (D7: 0.42). D1 covers only 3 outlets, so its $\sigma_k$ is an unstable conservative default; it carries negligible weight but is retained for completeness. D3 is our dual-PageRank score over the 2018--2019 \citet{cointet2021} web-link network (not Cointet's own Twitter-follower position), and is therefore stale for outlets that moved after 2019. The corrective is not D7 alone: a family-level decomposition (Appendix~\ref{appendix:shifters}) shows D7 contributes most at JDD but only fourth-most at CNews, where D0 and D6b carry more of the mass. Ownership (D0) is mapped through a discrete owner-structural lookup table (20 scored categories plus two null categories---for concealed ownership or an ideologically-unmarked owner, which receive null D0; the complete owner-category mapping is in the replication package). Because D0 is z-standardized, only the ordinal ranking of owner types matters (Communist Party (PCF) organ $<$ independent left $<$ public/cooperative $<$ conservative oligarch $<$ far-right aligned); see Table~\ref{tab:loo} for per-family LOO details.

\subsection{Per-Family Z-Standardization}

Each evidence family k is standardized independently across the subset of outlets with available data:
\[
z_{ik} = \frac{x_{ik} - \bar{x}_k}{s_k}
\]
where $\bar{x}_k$ and $s_k$ are the population mean and standard deviation computed over the n$_k$ outlets with non-null data in family k. Families with fewer than two observed values are set to zero (uninformative). Outlets missing family k are absent from that family's z-matrix entirely; they do not receive an imputed zero, which would introduce spurious shrinkage toward the center. For some families missingness is structural (D1 covers only specific outlet types), not random; since scores are never imputed, no missing-at-random assumption is needed.

This ensures comparability across heterogeneous scales (D2: 0--6 survey; D1: percentage points; D7: $\pm$1 ratio); per-family z-standardization is canonical in composite indicator construction \citep{oecd2008composite}.

\subsection{Latent Score via Empirical-Bayes Shrinkage}

FrIdéo is a composite-indicator measurement model that estimates a latent ideological position from multiple heterogeneous evidence families; unlike factor-analytic or ideal-point models, it does not infer positions through a generative statistical model.

For each outlet i, a blended latent score weights direct evidence against a historical editorial tradition prior: the more evidence families cover an outlet, the closer its score tracks their mean:

\begin{gather*}
\bar{z}^{\text{direct}}_i = \tfrac{1}{n_i} \sum_{k \text{ avail.}} z_{ik} \\
z^{\text{struct}}_i = 1.5\,\text{S2}_i, \qquad \alpha_i = \tfrac{n_i}{n_i + \lambda}\ (\lambda = 2) \\
z^{\text{blended}}_i = \alpha_i\, \bar{z}^{\text{direct}}_i + (1 - \alpha_i)\, z^{\text{struct}}_i
\end{gather*}

The shrinkage parameter $\lambda$ = 2 implies that an outlet with two direct families receives equal weight between empirical evidence and the structural prior ($\alpha$ = 0.50); four families yields $\alpha$ = 0.67; seven families yields $\alpha$ = 0.78. Outlets with no direct evidence (n = 0) fall back entirely on the structural prior. The structural prior S2 encodes founding ideological tradition on a symmetric five-point ordinal grid ($-$0.5 to +0.5), scaled by 1.5 to match the empirical z-score spread. S2 values were assigned by the authors from each outlet's documented founding profile---founding year, founder or institution, initial editorial alignment---using published press histories and editorial archives; the full per-outlet coding is in Appendix~\ref{appendix:s2}.

The prior acts as a regularizer pulling low-evidence outlets toward their documented historical identity, while high-evidence outlets are determined primarily by the direct pool. $\lambda$ = 2 is a design choice, not an estimate: it sets $\alpha$ = 0.50 at the minimum meaningful evidence count ($n_i$ = 2), encoding the judgment that two sources should not override the historical record without corroboration. We use \emph{empirical-Bayes} in the estimation sense that the family dispersions $\sigma_k$ are data-estimated (leave-one-family-out residual variance), while $\lambda$ and the prior mean (S2) are fixed by design; the procedure is thus empirical-Bayes-\emph{style}, not a fully estimated hierarchical model. Shrinkage depends on evidence quantity ($n_i$), not inter-family coherence, keeping point estimation and uncertainty quantification separate. $\bar{z}^{\text{direct}}_i$ is \emph{unweighted} rather than precision-weighted ($1/\sigma_k^2$): since $\sigma_k$ is largest for families that diverge from consensus (notably D7), precision weighting would suppress the contemporary signal the model is designed to keep auditable.

The blended scores are re-standardized across all 30 outlets:
\[
\text{final}_i = \frac{z^{\text{blended}}_i - \mu_B}{\sigma_B}
\]
where $\mu_B$ and $\sigma_B$ are the mean and population standard deviation of $\{z^{\text{blended}}_i\}$. The result is a zero-mean, unit-variance scale across the 30-outlet panel. No external anchors constrain this scale; the outlets at its extremes (L'Humanité on the left, Valeurs actuelles on the right) are validated post-hoc, not used to fix it.

\paragraph{The scale is panel-relative.} Because every family is z-standardized over the panel and the blended scores re-standardized across the same 30 outlets, the zero-point and unit are defined \emph{internally}; an anchor-free scale cannot also be addition-invariant. Adding or removing outlets therefore shifts the reference distribution and can move boundary-proximate outlets across $\sigma$-band thresholds even when rank-order is unchanged. FrIdéo scores are thus comparable only \emph{within} a fixed panel: applying the method to a different outlet set re-defines the distribution and requires re-estimation rather than transplanting absolute scores. Externally anchored calibration (Limitations) is left to future work.

Each outlet receives an analytic uncertainty estimate that grows with evidence sparsity:

\begin{gather*}
\text{var}_i = \sum_{k \text{ avail.}} \sigma_k^2 \\
\sigma^{\text{base}}_i = \frac{1}{\sigma_B} \cdot \frac{\sqrt{\text{var}_i}}{n_i} \\
\text{penalty}_i = \frac{\lambda}{n_i + \lambda} \\
\sigma_i = \sigma^{\text{base}}_i \cdot (1 + \text{penalty}_i)
\end{gather*}

The sparsity penalty ($\text{penalty}_i$) ensures that outlets relying heavily on the structural prior are assigned wider confidence intervals. The 95\% CI is $[\text{final}_i - 1.96\sigma_i,\; \text{final}_i + 1.96\sigma_i]$.

\paragraph{Covariance sensitivity.} The additive CI formula assumes family independence ($\text{Var}(\bar{z}_i) = n_i^{-2}\sum_k \sigma_k^2$). The three families most likely to co-vary are the externally-assessed families D2 (audience survey), D4 (bias rating), and D6b (third-party profiling); under a conservative pairwise $\rho$ = 0.3 among them, CI half-widths inflate by 3.1\% on average (max 12.1\%, at Libération, Le Monde and Le Figaro) and no outlet's label changes. Because the families are by design convergent (mean inter-family $r\!=\!0.84$, Section~\ref{sec:results}), the independence-based CIs should be read as lower bounds on uncertainty.

Seven ideology labels (\emph{far-left} through \emph{far-right}) are assigned using symmetric $\sigma$-band cuts at $\pm 1.5\sigma$, $\pm 0.75\sigma$, $\pm 0.25\sigma$ on the z-standardized scale, corresponding to the French seven-category political convention (\emph{extr\^{e}me gauche} `far-left' through \emph{extr\^{e}me droite} `far-right'). The nested widths are chosen so that every category is populated: under the left-skewed French press distribution, equal-width bins would leave the centre category degenerate. All 30 assignments match established characterizations in the literature; a sensitivity check confirms they are stable under alternative symmetric cut-point choices. For the three sparse outlets with a non-zero prior (Blast, Le HuffPost, Atlantico) that agreement is partly circular, the prior encoding the same historical characterization the label is checked against; refitting without the prior leaves their labels unchanged (Appendix~\ref{appendix:shifters}). Slate.fr and Challenges are not circular in this way---their prior is zero, so it exerts no directional pull---but face the converse problem of thin, unanchored evidence, and their labels should be read accordingly. Labels are provided for interpretability only; all substantive analyses use continuous scores. The unidimensional assumption underlying them is supported empirically (principal component analysis, PCA, on the four highest-coverage families; Section~\ref{sec:results} and Appendix~\ref{appendix:unidimensional}), though it may collapse distinct economic and cultural dimensions where these diverge (e.g.\ La Croix and Les Echos, both right-of-center via Catholic tradition vs.\ economic liberalism; see Limitations).

Because the blended score is a weighted sum, each family's signed contribution to the final score is exactly recoverable: $C_{ik} = (\alpha_i / n_i) \cdot z_{ik} / \sigma_B$ for direct families, and $C_{i,\text{prior}} = (1-\alpha_i) \cdot z^{\text{struct}}_i / \sigma_B$ for the structural prior. Appendix~\ref{appendix:decomp} reports these normalized additive contribution shares, grouped into five interpretable evidence classes, for all 30 outlets; absolute shares describe relative evidence weight, signed shares indicate directional pull.

\section{Content-Based Lexical Scoring (D7)}

D7 is computed from raw article text (headline + body) using regex term matching against 20 left-coded patterns (e.g., \emph{violences policières} `police violence', \emph{féminicide} `femicide', \emph{racisme systémique} `systemic racism', \emph{urgence climatique} `climate emergency') and 15 right-coded patterns (e.g., \emph{ensauvagement} `descent into savagery', \emph{grand remplacement} `great replacement', \emph{islamo-gauchisme} `Islamo-leftism', \emph{immigration massive/incontrôlée} `mass/uncontrolled immigration'). The lexicon's candidate pool was drawn from a three-month pilot corpus (\emph{politique} `politics'-tagged articles, July--September 2023) and curated manually against explicit criteria rather than by any automated correlation search; individual retain/reject decisions were validated against the 2024--2025 corpus and held-out MBFC labels; of roughly 80 candidates, we kept the 35 above and rejected the remainder for neutral saturation (e.g., \emph{délinquance} `delinquency', \emph{émeutiers} `rioters') or polysemy (e.g., \emph{séparatisme} `separatism'), with the full rejection log in Appendix~\ref{appendix:d7reject}.

An occurrence is a \emph{plain use} when its sentence contains none of D7's distancing signals and a \emph{qualified use} when any fires; absence defaults to plain use. D7 tests only for an explicit distancing signal, not endorsement.

The check applies to every occurrence; sentence boundary extends through any unclosed «\ldots» spanning a period break. Six signals qualify an occurrence: \emph{(i)~quotation enclosure}, the term falls inside French guillemets «…», ASCII/curly quotes, an unclosed «~within 150 chars, or a colon-plus-quote attribution; \emph{(ii)~distancing prefix}, ironic operators such as \emph{théorie du} `theory of', \emph{soi-disant} `so-called', \emph{prétendu} `alleged'; \emph{(iii)~side-specific critique vocabulary}, e.g.\ \emph{complotiste} `conspiracist', \emph{réfute} `refutes' for right-coded terms and \emph{supposé} `supposed', \emph{idéologie woke} `woke ideology' for left-coded (full lists in the replication package).

An article-level majority-vote rule applies symmetrically: an article is counted as a left- or right-signal only when plain-use occurrences for that side outnumber qualified-use occurrences.

\paragraph{LLM validation.} To validate the distancing classifier, we drew a stratified sample of 527 occurrences from the politique corpus (2023-H2 and 2024-H1, distinct from the all-category 2024--2025 scoring corpus), stratified by term side, D7 decision, and outlet ideology band, with outlet identity masked. Each occurrence was judged once by \texttt{gpt-oss-120b} \citep{openai2025gptoss}\footnote{117B MoE model (Apache 2.0), served via the Regolo API.} (temperature\,=\,0; prompt hash in the replication package) with a binary \emph{PLAIN USE} vs \emph{QUALIFIED USE} prompt following D7's rule: explicit textual signals only, reported speech excluded. Of the 527 drawn, 430 remain after stratified balancing, de-duplication, and removal of occurrences whose term did not survive lexicon curation. On these, D7 and the LLM agree at Cohen's $\kappa$ = 0.43 (and, computed separately on the left- and right-term subsets, $\kappa_{\text{left}}$ = 0.54, $\kappa_{\text{right}}$ = 0.31), with 118/430 disagreements (27.4\%).

The 312 agreement cases and the 118 adjudicated disagreements (each re-examined against the six signals) form the 430-occurrence gold standard. Against this gold, D7 achieves $\kappa$ = 0.938 (97.0\% agreement), outperforming the LLM judge ($\kappa$ = 0.49): D7 deterministically implements the six signals, whereas the LLM drifts into contextual reasoning even when instructed not to. On the 118 adjudicated rows alone, where the gold is not D7's own output, $\kappa$ = 0.78 (89.0\%). This validates the operational definition (the same rule applied unchanged to the 2024--2025 corpus), not the broader endorsement construct. As an independent stress-test, a logistic-regression probe on CamemBERT-base embeddings (768-d mean-pool, 5-fold CV) reaches macro F1 = 0.775 against D7's 0.969 on the same 430 occurrences, and an ensemble of the two underperforms D7 alone (F1 = 0.856): the 110M-parameter contextual model adds no measurable improvement on this operational task. Confusion matrices and UMAP visualisations are in Appendix~\ref{appendix:d7probe}. Adjudication was by a single author against a written codebook (released in the replication package). The headline $\kappa$ = 0.938 therefore measures fidelity to that codebook as applied by one coder, not agreement with an independently constituted gold standard; nor is the LLM comparison independent of it, since the adjudicator resolved the very disagreements the two produced. A second annotator (an unpaid volunteer, fluent in French, with no involvement in the D7 lexicon, rules or pipeline), working from the same codebook, independently re-labelled a blind, outlet-masked sample of 100 occurrences stratified over adjudicated and auto-gold cases, agreeing with the author's gold at $\kappa$ = 0.94 (boundary 0.94, control 0.93). On the 13 occurrences where the author's adjudication overruled the D7 rule, the second annotator reached the author's label in 12.

For each outlet across all sampled articles in a scoring period:
\[
\text{D7}_{\text{raw}} = \frac{n_{\text{right}} - n_{\text{left}}}{n_{\text{right}} + n_{\text{left}}}
\]
where $n_{\text{right}}$ is the count of articles for which plain-use right-term occurrences outnumber qualified-use ones, and $n_{\text{left}}$ the analogous count for left terms. A minimum of 10 signal articles is required; otherwise D7 is null for that outlet. Ten is the smallest count at which no single article dominates the majority-vote ratio (shifting D7\textsubscript{raw} by at most $0.2$ at $n=10$); it safeguards against sparse outlet samples. We score an all-category corpus pooled over 2024--2025 (568,906 sampled articles, capped at 1{,}000 per outlet-month); over this window every panel outlet clears the threshold (the thinnest, Mediapart, carries 102 signal articles), giving 30/30 coverage. We score all articles flagged by the curated political lexicon (Appendix~\ref{appendix:d7reject}) regardless of section tag, since \emph{politique} tagging is inconsistent across outlets. Spearman $\rho$ against the full-model z-scores is 0.93 ($n$ = 30); since D7 is itself a fused family, this is an internal-coherence check, not independent validation. Period stability is analyzed in Section~6.3, with per-outlet year-over-year shifts in the replication package.

The D7-only diagnostic submodel is reported in Appendix~\ref{appendix:d7}. Under the symmetric specification, CNews (+0.15) and Causeur (+0.38) remain clearly separated from Libération ($-$0.55) and L'Humanité ($-$0.83), and Le Monde scores center-left ($-$0.27), its residual undistanced security vocabulary outweighed by its left-coded usage.

\section{Results}
\label{sec:results}

Because editorial ideology has no observable ground truth, validation relies on convergent validity, robustness, and known-groups discrimination. The question is whether heterogeneous families induce a stable latent ordering that cleanly separates widely recognized ideological groups and persists under perturbation. We do not claim to have recovered the true position of any outlet; we claim a scale that is internally consistent, robust to modeling choices, and convergent with independent external evidence.

\paragraph{Cross-family convergence.} Pairwise correlations among the six best-covered families (D0, D3, D4, D6a, D6b, D7; $n\!\geq\!9$ per pair) range from $r\!=\!0.71$ to $r\!=\!0.94$ (mean $r\!=\!0.84$, 15 pairs), consistent with a single latent dimension (see Section~\ref{sec:framework}, Appendix~\ref{appendix:unidimensional}). A formal known-groups test confirms the separation: outlets the external MBFC taxonomy places on the left vs.\ right do not overlap on the no-D4 holdout scale, which excludes the grouping signal (AUC $=1.00$, Cohen's $d=3.05$, exact permutation $p=0.0006$; Appendix~\ref{appendix:knowngroups}).

\subsection{Outlet Ideology Scores}

The distribution spans $-$2.41 (L'Humanité) to +1.87 (Valeurs actuelles): 11 outlets score at or below center-left, 11 score at or above center-right, and 8 are centered. Fifteen outlets in the center region have 95\% CIs spanning zero and should be interpreted with caution; the five with $n\!\leq\!4$ direct evidence families (Blast, Le HuffPost, Slate.fr, Challenges, Atlantico) are the most heavily shrunk toward the prior (for Slate.fr and Challenges, whose prior is zero, toward the panel center; see Appendix~\ref{appendix:decomp} for contribution-share diagnostics).

Figure~\ref{fig:spectrum} displays the full ordering with 95\% confidence intervals; complete numerical scores are in Appendix Table~\ref{tab:fullmodel}.

\begin{figure*}[t]
\centering
\includegraphics[width=\textwidth]{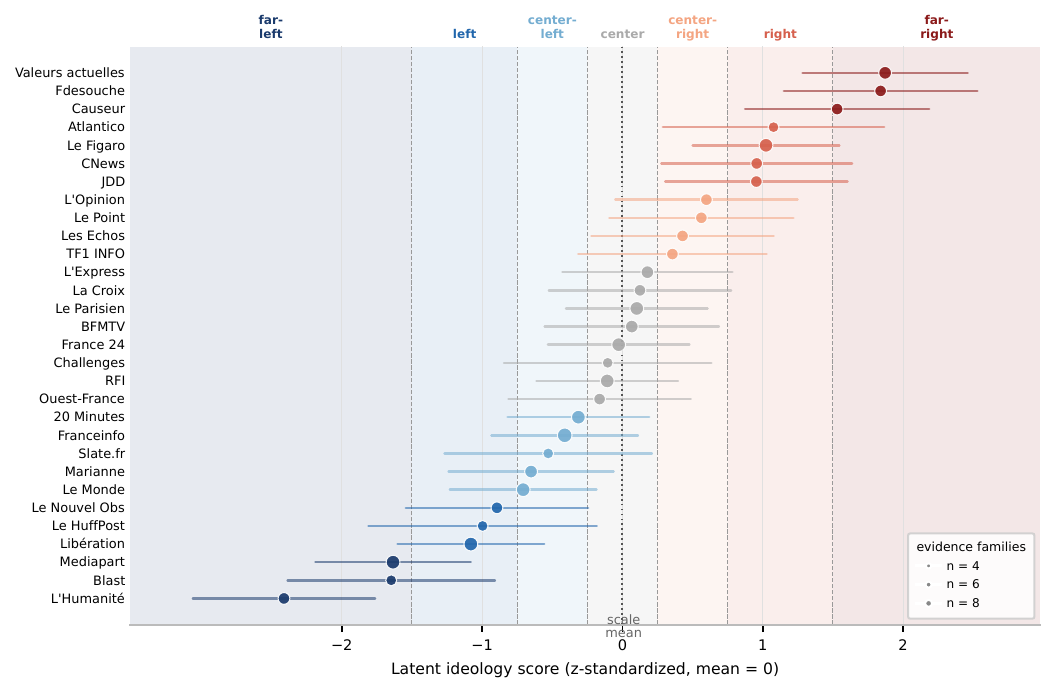}
\caption{FrIdéo latent ideology scores for 30 French news outlets. Points are full-model continuous scores (z-standardized); whiskers are 95\% CIs. Dot size encodes direct evidence families (range: 4--8). Band shading and dashed lines mark $\sigma$-band cut-points for the seven interpretive labels (orientation only; all substantive analyses use continuous scores).}
\label{fig:spectrum}
\end{figure*}

The center-left cluster (20 Minutes through Le Monde, $-$0.31 to $-$0.71) is denser than the right tail, a shape that converges with independent surveys of the French press (though partly a function of panel composition; see Limitations). CNews (+0.96) and Le Figaro (+1.02) are effectively tied within the right cluster, with heavily overlapping CIs, consistent with CNews's rightward shift following Bolloré's 2015 takeover of its parent network and the subsequent 2017 iTélé--CNews relaunch \citep{cage2024}. At the far-right pole, Valeurs actuelles (+1.87) and Fdesouche (+1.84) are nearly indistinguishable: both carry uniformly strong rightward evidence, and VA edges ahead only on its additional ownership family (D0\,$=\,+$2.13$\sigma$; Fdesouche's anonymous ownership yields null D0). Causeur (+1.53) enters the far-right band only at the margin (barely above the $+$1.5$\sigma$ cut-point), polemical but less uniformly extreme than the two identitarian outlets.

\subsection{Secondary External Corroboration: MBFC Holdout}

We hold out the Media Bias Fact Check (D4) evidence family and repurpose it as a secondary external corroboration check, not as the criterion the model is intended to reconstruct. This no-D4 holdout model achieves Spearman $\rho$ = 0.918 (95\% CI [0.77,\,0.97]; $p < 0.001$) against MBFC labels (n = 15). Partial correlation among D4, D2, and D6b is expected under convergent validity \citep{campbellFiske1959}; to rule out circularity, we also compare MBFC against the five families with no methodological overlap (D0, D1, D3, D6a, D7). This structurally independent model agrees at $\rho$ = 0.931, essentially matching the full-model value, so the MBFC-overlapping families do not drive agreement. Removing D4 shifts continuous scores at $\rho$ = 0.9969 (n = 30); maximum shift is 0.224$\sigma$ (Le HuffPost). Two outlets near $\sigma$-band cut-points cross a label boundary (Le Monde, 20 Minutes); continuous scores are preferred downstream.

Ordinal agreement (Table~\ref{tab:mbfc}): exact match 6/15, adjacent 8/15, distant 1/15. The sole distant case (Mediapart) reflects a \emph{content--structure gap}: its network position and reader-funded identity align far-left (no-D4 score $-$1.71) while MBFC's content-based rater registers its sober investigative style as less partisan (Left-Center).

\subsection{Sensitivity Analysis}

We drop each of the nine direct families in turn, rerun the full pipeline, and compute Spearman $\rho$ and maximum absolute score shift against the full-model scores (Table~\ref{tab:loo}). The lowest leave-one-out rank correlation is $\rho$ = 0.992 (dropping D6a); the largest single absolute score shift is 0.27$\sigma$ (dropping D3, the Cointet network family, at Le Figaro). Dropping D7, the only within-period text family and thus structurally orthogonal to the historical evidence pool, shifts scores by at most 0.25$\sigma$ (at JDD) yet leaves rank order essentially intact. Across all nine families, LOO $\rho$ $\geq$ 0.992; no single family is load-bearing.

Bootstrap stability across 10,000 joint perturbations of all $\sigma_k$ and $\lambda \in [1.0, 3.0]$ is $\rho$ = 0.996 (mean), min 0.988; under uniform $\sigma_k$ = 0.80 defaults ranks are unchanged ($\rho$ > 0.999). Deterministic refits at $\lambda = 1$ and $\lambda = 3$ move every score $<0.08\sigma$ ($\rho \geq 0.996$); the sole label change is L'Express, which crosses the center/center-right boundary at $\lambda = 1$ but not $\lambda = 3$.

The distancing layers consistently sharpen the left--right separation (Table~\ref{tab:loo}, right panel). The guillemet filter is the single largest step and acts asymmetrically: it moves left outlets further left (L'Humanité: $-$0.678 → $-$0.822; Libération: $-$0.394 → $-$0.545), because they cite right-coded vocabulary critically (in reported speech), which bag-of-words (BoW) scoring miscounts as rightward; the prefix and symmetric layers add incremental shifts. Right-leaning outlets move only slightly (CNews $+$0.225 → $+$0.145; Causeur $+$0.342 → $+$0.423): their right-coded vocabulary appears predominantly as plain use, so little is filtered. On the all-category corpus the left outlets are already net-negative under BoW: their broad output carries enough plain left-coded usage to set the sign, so the distancing rule sharpens rather than reverses them; its corrective value is largest precisely where right-coded vocabulary is cited critically.

No single term block is determinative: dropping any one of the ten D7 term blocks (the 35 lexical patterns are grouped into five left and five right semantic blocks for this ablation) leaves the fused rank order largely intact ($\rho \geq 0.89$ vs.\ the full model). The 10-article threshold still matters for narrow sub-periods (e.g.\ the JDD experiment), where a single half-year can fall below it.

D7 is stable year-over-year when the editorial line is stable, yet moves sharply when it shifts (JDD experiment below). Scoring the all-category corpus separately by calendar year yields $\rho$ = 0.95 between the 2024 and 2025 D7 scores (n = 30; mean $|\Delta|$ = 0.09; per-outlet shifts in the replication package), comparable to the static families. The largest year-over-year shifts are modest ($|\Delta|\leq 0.27$, at Ouest-France) and reflect ordinary corpus variation rather than editorial change; no outlet crosses a label boundary. Contrasted with the sharp JDD discontinuity (Section~\ref{sec:discussion}), this confirms D7 moves with genuine editorial change, not sampling noise.

\section{Discussion}
\label{sec:discussion}

D7 is an imperfect standalone: security-policy paraphrase (e.g., \emph{violences urbaines} `urban violence' without guillemets) registers identically to plain use when no distancing signal fires, so a center-left outlet can accrue a residual right-coded offset from descriptive usage. D7 is a family within the fusion, not a sufficient statistic.

\paragraph{The JDD natural experiment.} The Bolloré overhaul of 22 June 2023 (installing Geoffroy Lejeune, formerly editor of \emph{Valeurs actuelles}, as editorial director in place of Jérôme Béglé) provides a natural experiment with clean before/after structure. D7\textsubscript{raw} traces a monotone rightward trajectory tightly aligned with the event (Table~\ref{tab:jdd}): net \emph{left} pre-overhaul (consistent with JDD's establishment identity), then a $+$0.46 swing in the first six months that flips it to clearly right, climbing further by 2024-H1; all three windows use the identical scoring procedure, so the shift reflects editorial content, not a scoring change. Post-overhaul, JDD treats \emph{immigration massive} (`mass immigration') and \emph{ensauvagement} (`descent into savagery') as descriptive rather than contested vocabulary, which the plain/qualified rule registers directly. The discontinuity is the paper's sharpest demonstration of D7's distinctive property: none of the eight remaining families can register the transition (ownership, citation networks, historical ratings are static within period), whereas D7 detects it temporally.

\begin{table}[h!]
\begin{nolinenumbers}
\centering
\small
\setlength{\tabcolsep}{6pt}
\caption{JDD D7\textsubscript{raw} trajectory across the 22 June 2023 Bolloré overhaul: the plain-use majority shifts monotonically rightward.}
\label{tab:jdd}
\begin{tabularx}{\columnwidth}{@{}l >{\RaggedRight\arraybackslash}Xr@{}}
\toprule
Period & Relation to overhaul & D7\textsubscript{raw} \\
\midrule
Pre-overhaul    & Jan 2022 -- Jun 2023 & $-$0.160 \\
2023-H2         & immediately after    & $+$0.298 \\
2024-H1         & $\sim$1 year after   & $+$0.573 \\
\bottomrule
\end{tabularx}
\end{nolinenumbers}
\end{table}

\section{Conclusion}

We presented FrIdéo, a continuous, ecosystem-wide ideology scale for 30 French national news outlets, fusing nine heterogeneous evidence families by per-family z-standardization and empirical-Bayes shrinkage toward a founding-tradition prior. Its text family, D7, introduces symmetric plain/qualified scoring, separating plain use of ideologically loaded vocabulary from distanced use on both sides and so reading editorial position from current articles. This makes the scale time-sensitive: D7 registers JDD's post-2023 realignment, which the eight static families cannot. The ordering survives dropping any single family (LOO $\rho \geq 0.992$) and joint perturbation (bootstrap $\rho$ = 0.996), and agrees with held-out MBFC labels. We release the scores, confidence intervals, and replication code.

\section*{Limitations}

Our source-consistency approach treats disagreement with the remaining evidence pool as a proxy for noise. That assumption is plausible for this applied fusion setting but not guaranteed: a family that captures genuine recent editorial change will disagree with historical consensus sources precisely because it is more informative about current position. Users who need maximum temporal sensitivity for a specific recently-shifted outlet should weight D7 more heavily or re-run the model with conservative $\sigma_k$ assignments (tooling provided in the replication package).

The evidence families span 2018 \citep{pew2018} to 2026 \citep{mbfc2026}, and French media has undergone significant ownership concentration and editorial drift in that period. CNews's post-2017 Bolloré transformation and Marianne's ongoing nationalist pivot are captured imperfectly by the historical families. For outlets undergoing rapid shift within the scoring window (as the JDD analysis demonstrates), D7 should be weighted more heavily or re-run on targeted sub-periods. Missingness handling (Section~4.2) is most defensible for D0 (anonymous or contested ownership) and least so for D6b (Wikipedia profiling), where null could reflect insufficient notability, potentially correlated with outlet age or audience size.

FrIdéo assumes a unidimensional left-right ideology space, consistent with the outlet-scoring literature \citep{gentzkowShapiro2010,budak2016}; PCA on the four highest-coverage families is consistent with a single dominant dimension, including a La Croix/Les Echos test that finds little evidence of a separable economic-vs-cultural second axis (Appendix~\ref{appendix:unidimensional}). Extending FrIdéo to a multidimensional framework (the economic/cultural collapse flagged in Section~\ref{sec:framework}) is a natural direction for future work; candidate subdimension loadings include D0 (ownership → economic axis) and D6b (third-party profiling → cultural axis). The D7 lexical inventory was manually curated using explicit inclusion/exclusion criteria (Appendix~\ref{appendix:d7reject}) and remains a source of researcher degrees of freedom that can affect scores for sparsely attested outlets (Blast, Slate.fr). Event-driven but ideologically neutral vocabulary is a particular hazard: \emph{émeutiers} (``rioters''), for example, is excluded for neutral saturation, its removal confirmed on the scoring corpus and held-out MBFC (it fires on factual reporting such as the May 2024 Nouvelle-Calédonie unrest rather than on editorial stance). The full-model shrinkage further absorbs residual D7 noise via evidence-pool consensus. Finally, the left-denser shape of the distribution is partly a function of panel composition: a different outlet set would re-define the reference distribution, so we read the distributional shape as convergent with, rather than confirmatory of, independent surveys. An externally anchored calibration, for example projecting the media scores onto the Chapel Hill Expert Survey (CHES) party-placement scale, would yield an addition-invariant, cross-panel-comparable scale and would quantify the systematic rightward offset of the media centre relative to the party spectrum. We regard this as the primary planned extension but leave it to future work, since it requires an external bridge between party and media placements that the present evidence families do not supply.

\section*{Reproducibility Statement}

The core scoring pipeline (z-standardization, empirical-Bayes shrinkage, bootstrap, LOO) depends only on the Python standard library; PCA (unidimensionality validation) and the optional GMM label-assignment path use \texttt{scikit-learn}, while the D7 lexical scoring, shrinkage estimation, and all sensitivity analyses require no external ML framework. The bootstrap random seed is fixed at 42. Full replication code, outlet evidence data, scored results, and a pinned dependency lockfile are available at \href{https://github.com/lefrenchnewslab/frideo}{\mbox{\nolinkurl{https://github.com/}}\allowbreak\mbox{\nolinkurl{lefrenchnewslab/frideo}}}.

\bibliography{../frideo_refs}

\appendix
\numberwithin{figure}{section}
\numberwithin{table}{section}
\setcounter{figure}{0}
\setcounter{table}{0}

\section{Evidence Families}
\label{appendix:families}

Table~\ref{tab:families} specifies the ten evidence inputs of Section~\ref{sec:framework}---source, construct measured, source-consistency parameter $\sigma_k$, and outlet coverage. Coverage ranges by design from 3/30 (D1, TV only) to 30/30 (D3, D7, S2); it is this variation the shrinkage step absorbs.

\begin{table*}[tp]
\begin{nolinenumbers}
\centering
\scriptsize
\setlength{\tabcolsep}{3pt}
\caption{Evidence families used in the ideology scoring framework.}
\label{tab:families}
\begin{tabularx}{\textwidth}{@{}l
  >{\hsize=0.95\hsize\RaggedRight\arraybackslash}X
  >{\hsize=1.35\hsize\RaggedRight\arraybackslash}X
  >{\hsize=0.70\hsize\centering\arraybackslash}X
  >{\hsize=1.00\hsize\RaggedRight\arraybackslash}X@{}}
\toprule
ID & Source & What it measures & Source consistency parameter $\sigma_k$ & Outlet coverage \\
\midrule
D0 & Acrimed/\emph{Le Monde diplo.} ownership map (via data.gouv.fr); press/Wikipedia for absent or recently-sold outlets & Owner political ideology via structural map & 0.35 & 28/30 (null: concealed or ideologically-unmarked owner) \\
D1 & \citet{cage2022} & Guest radical-right deviation relative to France 2 benchmark (pp) & 0.57 & 3/30 (TV channels) \\
D2 & \citet{pew2018} & Audience-perceived ideology (0--6 scale; 3 = center), directly-published placement means & 0.57 & 8/30 \\
D3 & \citet{cointet2021} & Our continuous position via dual PageRank over Cointet's web-link network (421 nodes, 20,992 edges), seeded on its community blocks; right/left log-ratio & 0.63 & 30/30 (28 PPR + 2 categorical fallback) \\
D4 & \citet{mbfc2026} & Categorical bias label (far-left to far-right) & 0.50 & 15/30 \\
D6a & Editorial charters / statutes & Self-declared political orientation & 0.48 & 18/30 \\
D6b & Wikipedia / press typology & Third-party journalistic profiling & 0.30 & 27/30 \\
D7 & This work (signal-based lexical) & Content-based lexical ideology: plain-use occurrences outnumber qualified-use (signal-present) occurrences & 0.42 & 30/30 \\
D8 & Ad Fontes Media & Categorical bias label from independent US rater (same ordinal scale as D4; normalised via D4 reference distribution) & 0.24 & 9/30 \\
S2 & Historical record & Founding editorial tradition (5-point scale) & prior only & 30/30 \\
\bottomrule
\end{tabularx}
\end{nolinenumbers}
\end{table*}

\section{Sensitivity and Distancing-Ablation Tables}
\label{appendix:sensitivity}

Table~\ref{tab:loo} supports Section~6.3. Its left panel reports the leave-one-family-out analysis (drop each direct family, refit, compare to the full model); its right panel isolates each distancing layer in D7 across five high-signal outlets, from raw bag-of-words to the full symmetric check.

\begin{table*}[t]
\begin{nolinenumbers}
\centering
\small
\setlength{\tabcolsep}{3pt}
\caption{\textbf{Left:} LOO sensitivity --- rank correlation and maximum absolute score shift when each direct family is dropped in turn. \textbf{Right:} D7 distancing ablation for five high-signal outlets (2024--2025); BoW = raw bag-of-words, Guill.\ = guillemet filter, G+P = adds distancing prefix, Full = full symmetric (adds side-specific critique markers); Full differs from Table~\ref{tab:d7full} by $<$0.05 (per-year vs.\ pooled).}
\label{tab:loo}
\begin{tabularx}{\textwidth}{@{}>{\RaggedRight\arraybackslash}Xrrr>{\RaggedRight\arraybackslash}X | >{\RaggedRight\arraybackslash}Xrrrr@{}}
\toprule
Family & $n$ & LOO $\rho$ & Max $|\Delta|$ & Outlet & Outlet & BoW & Guill. & G+P & Full \\
\midrule
D0 (ownership) & 28 & 0.9973 & 0.196 & Libération & L'Humanité & $-$0.678 & $-$0.822 & $-$0.830 & $-$0.836 \\
D1 (Cagé guests) & 3 & 0.9969 & 0.145 & Le Figaro & Libération & $-$0.394 & $-$0.545 & $-$0.547 & $-$0.550 \\
D2 (Pew survey) & 8 & 0.9973 & 0.182 & Le Monde & Le Monde & $-$0.196 & $-$0.297 & $-$0.290 & $-$0.289 \\
D3 (Cointet PPR) & 30 & 0.9956 & 0.268 & Le Figaro & CNews & +0.225 & +0.149 & +0.147 & +0.145 \\
D4 (MBFC) & 15 & 0.9969 & 0.224 & Le HuffPost & Causeur & +0.342 & +0.410 & +0.416 & +0.423 \\
D6a (editorial charters) & 18 & 0.9920 & 0.207 & JDD &  &  &  &  &  \\
D6b (Wikipedia profiling) & 27 & 0.9969 & 0.165 & Le Figaro &  &  &  &  &  \\
D7 (signal-based lexical) & 30 & 0.9933 & 0.250 & JDD &  &  &  &  &  \\
D8 (Ad Fontes Media) & 9 & 1.0000 & 0.059 & TF1 INFO &  &  &  &  &  \\
\bottomrule
\end{tabularx}
\end{nolinenumbers}
\end{table*}

\section{Full-Model Outlet Ideology Scores}
\label{appendix:scores}

Table~\ref{tab:fullmodel} is the released resource: final score, analytic standard error, 95\% CI, seven-band label, and direct-family count for all 30 outlets. Scores are zero-mean and unit-variance across the panel by construction (Section~4.3), hence interpretable only relative to this outlet set.

\begin{table*}[t]
\begin{nolinenumbers}
\centering
\footnotesize
\setlength{\tabcolsep}{3pt}
\caption{Full-model ideology scores, all 30 outlets, two panels. Label is for interpretability only; all analyses use the continuous Score. The analytic $\sigma_i$ is recoverable to within 0.01 as one-quarter of the CI width divided by 0.98. $^{\dagger}$CI spans zero (15 outlets, all near-center). Five outlets have $n\!\leq\!4$ direct families (Blast, Le HuffPost, Slate.fr, Challenges, Atlantico): shrinkage places roughly a third of the weight on the tradition prior ($\alpha = 0.67$ at $n=4$), so these are heavily shrunk and their labels indicative only. For Slate.fr and Challenges the prior is zero (S2 = 0), contributing no signed mass; shrinkage instead pulls them toward the panel center. See Appendix~\ref{appendix:decomp}.}
\label{tab:fullmodel}
\begin{tabularx}{\textwidth}{@{}>{\RaggedRight\arraybackslash}Xr>{\RaggedRight\arraybackslash}X>{\RaggedRight\arraybackslash}Xr | >{\RaggedRight\arraybackslash}Xr>{\RaggedRight\arraybackslash}X>{\RaggedRight\arraybackslash}Xr@{}}
\toprule
Outlet & Score & 95\% CI & Label & $n$ & Outlet & Score & 95\% CI & Label & $n$ \\
\midrule
L'Humanité & $-$2.41 & [$-$3.06, $-$1.76] & far-left & 5 & BFMTV$^{\dagger}$ & +0.07 & [$-$0.55, +0.69] & center & 6 \\
Blast & $-$1.65 & [$-$2.39, $-$0.91] & far-left & 4 & Le Parisien$^{\dagger}$ & +0.10 & [$-$0.40, +0.61] & center & 7 \\
Mediapart & $-$1.63 & [$-$2.19, $-$1.08] & far-left & 7 & La Croix$^{\dagger}$ & +0.13 & [$-$0.52, +0.77] & center & 5 \\
Libération & $-$1.08 & [$-$1.60, $-$0.56] & left & 7 & L'Express$^{\dagger}$ & +0.18 & [$-$0.43, +0.78] & center & 6 \\
Le HuffPost & $-$1.00 & [$-$1.81, $-$0.18] & left & 4 & TF1 INFO$^{\dagger}$ & +0.36 & [$-$0.31, +1.03] & center-right & 5 \\
Le Nouvel Obs & $-$0.89 & [$-$1.54, $-$0.24] & left & 5 & Les Echos$^{\dagger}$ & +0.43 & [$-$0.22, +1.08] & center-right & 5 \\
Le Monde & $-$0.71 & [$-$1.23, $-$0.18] & center-left & 7 & Le Point$^{\dagger}$ & +0.56 & [$-$0.09, +1.22] & center-right & 5 \\
Marianne & $-$0.65 & [$-$1.24, $-$0.06] & center-left & 6 & L'Opinion$^{\dagger}$ & +0.60 & [$-$0.05, +1.25] & center-right & 5 \\
Slate.fr$^{\dagger}$ & $-$0.53 & [$-$1.27, +0.21] & center-left & 4 & JDD & +0.95 & [+0.31, +1.60] & right & 5 \\
Franceinfo$^{\dagger}$ & $-$0.41 & [$-$0.93, +0.11] & center-left & 8 & CNews & +0.96 & [+0.28, +1.64] & right & 5 \\
20 Minutes$^{\dagger}$ & $-$0.31 & [$-$0.82, +0.19] & center-left & 7 & Le Figaro & +1.02 & [+0.50, +1.55] & right & 7 \\
Ouest-France$^{\dagger}$ & $-$0.16 & [$-$0.81, +0.49] & center & 5 & Atlantico & +1.08 & [+0.29, +1.87] & right & 4 \\
RFI$^{\dagger}$ & $-$0.11 & [$-$0.61, +0.40] & center & 7 & Causeur & +1.53 & [+0.87, +2.19] & far-right & 5 \\
Challenges$^{\dagger}$ & $-$0.10 & [$-$0.84, +0.63] & center & 4 & Fdesouche & +1.84 & [+1.15, +2.53] & far-right & 5 \\
France 24$^{\dagger}$ & $-$0.03 & [$-$0.53, +0.48] & center & 7 & Valeurs actuelles & +1.87 & [+1.28, +2.46] & far-right & 6 \\
\bottomrule
\end{tabularx}
\end{nolinenumbers}
\end{table*}

\section{Normalized Additive Contribution Shares}
\label{appendix:decomp}

\begin{figure*}[t]
\begin{nolinenumbers}
\centering
\includegraphics[width=\textwidth]{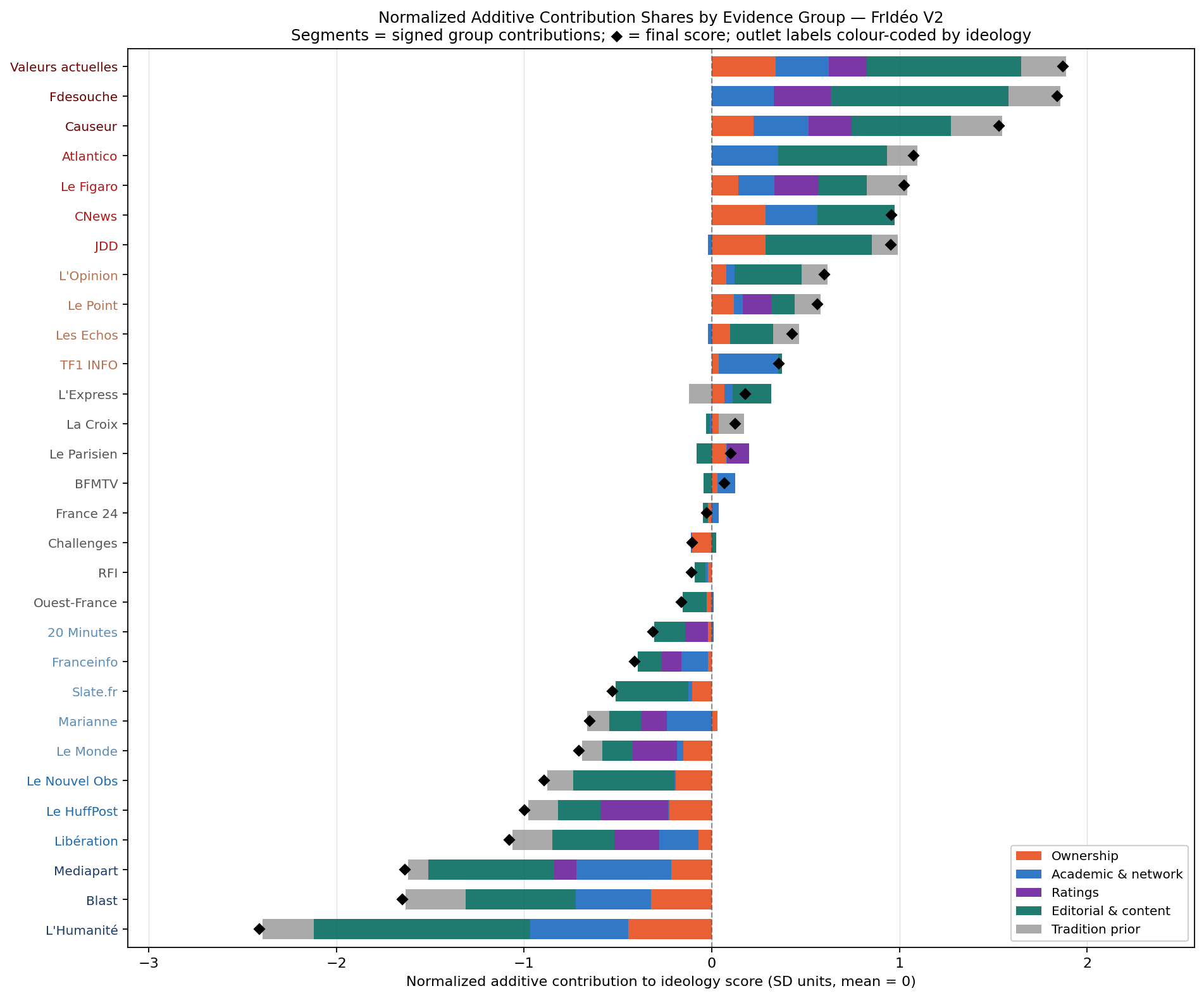}
\caption{Normalized additive contribution shares by evidence group, all 30 outlets (spectrum left to right, bottom to top). Each segment sums signed final-scale contributions ($C_{ig}$, SD units) within a group; the diamond~($\blacklozenge$) marks the final score. Groups: \textbf{orange} Ownership (D0); \textbf{blue} Academic \& network (D1--D3); \textbf{purple} Ratings (D4, D8); \textbf{teal} Editorial \& content (D6a, D6b, D7); \textbf{grey} Tradition prior (S2).}
\label{fig:decomp}
\end{nolinenumbers}
\end{figure*}

Figure~\ref{fig:decomp} decomposes each final score into signed additive contributions from five evidence groups, defined by methodological provenance: Ownership (D0), Academic \& network (D1, D2, D3), Ratings (D4, D8), Editorial \& content (D6a, D6b, D7), and the Tradition prior (S2). For outlet $i$, group $g$ contributes $C_{ig} = \sum_{k \in g} C_{ik}$, with $C_{ik} = (\alpha_i / n_i) \cdot z_{ik} / \sigma_B$ for direct families and $C_{i,\text{prior}} = (1-\alpha_i) \cdot z^{\text{struct}}_i / \sigma_B$ for the prior, on the final standardized scale (SD units). These are \emph{normalized additive decomposition shares}, not causal attributions or Shapley values.

Three patterns emerge. Evidence-rich outlets (Le Monde $n=7$, Franceinfo $n=8$, Valeurs actuelles $n=6$) show balanced multi-group support, the prior contributing under 25\% of absolute signal mass. Sparse outlets (Blast, Slate.fr, both $n=4$) are dominated by at most two groups, with roughly a third of shrinkage weight ($1-\alpha$) on the prior; where S2 = 0 (Slate.fr) it contributes none and shrinkage pulls toward the panel center. For near-center outlets (BFMTV, Ouest-France) positive and negative contributions offset: a small score reflects genuine evidence tension, not absent signal. Full family-level decomposition: replication package.

\section{Post-2019 Shifters and the No-Prior Refit}
\label{appendix:shifters}

\paragraph{Which family corrects a stale D3?} Decomposing the final score into signed family contributions (Appendix~\ref{appendix:decomp}) gives, in SD units and \% of absolute mass: \textbf{JDD} --- D7 $+$0.355 (32\%), D0 $+$0.286 (26\%), D6b $+$0.260 (24\%); \textbf{CNews} --- D0 $+$0.286 (29\%), D6b $+$0.260 (27\%), D1 $+$0.241 (25\%), D7 $+$0.152 (16\%). D7 leads at the ownership change it was designed to catch, but is not the general corrective.

This is also why LOO understates D7: dropping it moves JDD only 0.25$\sigma$ (Table~\ref{tab:loo}) despite supplying 32\% of its mass, because D0, D6b and the prior pull the same way. LOO measures marginal displacement under corroboration, not share of evidence---D6a's largest shift is likewise at JDD (0.207).

\paragraph{No-prior refit.} Refitting with $\alpha_i = 1$ throughout (prior removed, then re-standardized) tests whether the face-validity agreement is circular. All three prior-dependent sparse outlets keep their labels: Blast $-$1.65 $\rightarrow$ $-$1.65, Le HuffPost $-$1.00 $\rightarrow$ $-$1.04, Atlantico $+$1.08 $\rightarrow$ $+$1.16. Only two labels change panel-wide, neither sparse: L'Express (center $\rightarrow$ center-right) and Causeur (far-right $\rightarrow$ right), both already flagged as cut-point sensitive.

\section{Validation}
\label{appendix:validation}

This appendix collects the three validity checks of Section~\ref{sec:results}: convergent agreement with a held-out rating source (\ref{appendix:mbfc}), separation of externally-defined known groups (\ref{appendix:knowngroups}), and a test of the unidimensionality assumption (\ref{appendix:unidimensional}). All three use holdout or out-of-model quantities, so the criterion is never also an input.

\subsection{MBFC Holdout: Outlet-Level Comparison}
\label{appendix:mbfc}

Table~\ref{tab:mbfc} compares labels from the \emph{no-D4 holdout model}---refitted with the MBFC family removed---against MBFC's own labels for the 15 outlets it covers. Since D4 is excluded from the model being evaluated, agreement is convergent corroboration, not reconstruction of a fitted target.

\begin{table}[tbp]
\begin{nolinenumbers}
\centering
\small
\setlength{\tabcolsep}{3pt}
\caption{No-D4 convergent validity vs.\ MBFC. Model = \emph{no-D4 holdout} labels, not full-model labels (e.g.\ Le Monde is center-left in the full model, left here).}
\label{tab:mbfc}
\begin{tabularx}{\columnwidth}{@{}>{\RaggedRight\arraybackslash}Xr>{\RaggedRight\arraybackslash}X>{\RaggedRight\arraybackslash}X>{\RaggedRight\arraybackslash}X@{}}
\toprule
Outlet & Score & Model & MBFC & Agree \\
\midrule
Mediapart & $-$1.71 & far-left & Left-Center & $\times$ (2 cat.) \\
Libération & $-$1.20 & left & Left-Center & $\sim$ (adjacent) \\
Le Monde & $-$0.79 & left & Left-Center & $\sim$ (adjacent) \\
Le HuffPost & $-$0.77 & left & Far Left & $\sim$ (adjacent) \\
Marianne & $-$0.60 & center-left & Left-Center & $\checkmark$ \\
Franceinfo & $-$0.35 & center-left & Left-Center & $\checkmark$ \\
20 Minutes & $-$0.20 & center & Left-Center & $\sim$ (adjacent) \\
RFI & $-$0.10 & center & Center/Least Biased & $\checkmark$ \\
France 24 & $-$0.01 & center & Center/Least Biased & $\checkmark$ \\
Le Parisien & +0.00 & center & Right-Center & $\sim$ (adjacent) \\
Le Point & +0.47 & center-right & Right-Center & $\checkmark$ \\
Le Figaro & +1.18 & right & Right-Center & $\sim$ (adjacent) \\
Causeur & +1.50 & far-right & Right & $\sim$ (adjacent) \\
Fdesouche & +1.77 & far-right & Far Right & $\checkmark$ \\
Valeurs actuelles & +1.89 & far-right & Right & $\sim$ (adjacent) \\
\bottomrule
\end{tabularx}
\end{nolinenumbers}
\end{table}

\subsection{Known-Groups Separation}
\label{appendix:knowngroups}

Do outlets an \emph{external} taxonomy assigns left vs.\ right separate on the FrIdéo scale? Groups follow MBFC bands (left = \emph{Far Left}/\emph{Left-Center}; right = \emph{Right-Center}/\emph{Right}/\emph{Far Right}; the two \emph{Center} outlets are reported but excluded). Scores are the \emph{no-D4 holdout} scores, so the grouping variable is not an input to what it separates. Separation is complete: every right-group outlet outscores every left-group outlet. The left group ($n=7$) averages $-$0.80 (range $-$1.71 to $-$0.20), the right ($n=6$) $+$1.14 ($+$0.00 to $+$1.89), the two Center outlets $-$0.05. Mann--Whitney $U=42/42$ (AUC $=1.00$), Cohen's $d=3.05$, exact one-sided permutation $p=0.0006$ ($1/1716$).

\subsection{Unidimensionality Validation}
\label{appendix:unidimensional}

We test by PCA whether the space is unidimensional and whether a second dimension is economic-versus-cultural. Inputs are the four highest-coverage families (D0, D3, D6b, D7; each $\geq$27 outlets) over the 25 complete-case outlets, z-standardized as in the main pipeline, eigen-decomposed without rotation.

\paragraph{Results.} PC1 explains 88.5\% of shared variance, PC2 7.4\% (scree ratio 11.9$\times$). All four families load strongly on PC1 (D0 $-$1.02, D6b $-$0.97, D7 $-$0.95, D3 $-$0.85), consistent with one dominant latent dimension. PC2 is driven mainly by D3 ($-$0.49), the rest loading weakly ($|$loading$|\leq$0.18) --- little evidence the residual axis is economic-versus-cultural. Scree plot and biplot: replication package.

\paragraph{Economic vs.\ cultural test.} If PC2 separated economic conservatism (D0) from cultural conservatism (D6b), La Croix and Les Echos --- which reach right-of-center by different routes --- should sit at opposite PC2 poles. They differ by only 0.35 SD (La Croix $-$0.10, Les Echos $+$0.25), and D3 loads opposite to D0 on PC2 ($-$0.49 vs.\ $+$0.14): network position and ownership capture partially distinct signals, not a clean economic/cultural split.

\section{Structural Prior (S2) Coding}
\label{appendix:s2}

S2 is the prior, not a direct family. It encodes \emph{founding} editorial tradition only --- deliberately distinct from current ownership (D0) and current self-declaration or profiling (D6a/D6b) --- so it reflects historical baseline, not present-day line. Authors assigned values on a symmetric five-point grid \{$-$0.5, $-$0.25, 0, $+$0.25, $+$0.5\} from each outlet's documented founding profile (year, founder or institution, initial alignment) using published press histories; per-outlet sources are in the replication package. The $\pm$0.5 cap is by design: the prior is weak relative to direct evidence and is down-weighted by shrinkage ($1-\alpha_i$) as families accrue, so it anchors only data-sparse outlets and is overridden wherever direct families exist (JDD's $+$0.25 founding tradition is dominated by its post-2023 D6b signal). Coverage 30/30.

\begin{table*}[t]
\begin{nolinenumbers}
\centering
\footnotesize
\setlength{\tabcolsep}{3pt}
\caption{Complete S2 founding-tradition coding for all 30 outlets, in two panels. Value is the ordinal S2 input (scaled by 1.5 in the model). Coverage 30/30.}
\label{tab:s2}
\begin{tabularx}{\textwidth}{@{}l c >{\RaggedRight\arraybackslash}X | l c >{\RaggedRight\arraybackslash}X@{}}
\toprule
Outlet & S2 & Founding basis & Outlet & S2 & Founding basis \\
\midrule
L'Humanité & $-$0.5 & 1904, Jaurès --- PCF organ & Ouest-France & 0.0 & 1944 (SIPA) --- independent \\
Blast & $-$0.5 & 2021, D.\ Robert --- left investigative & RFI & 0.0 & 1975 --- public international radio \\
Libération & $-$0.5 & 1973, Sartre --- radical left & France 24 & 0.0 & 2006 --- public international news \\
Mediapart & $-$0.25 & 2008, Plenel --- investigative, left & Challenges & 0.0 & 1982 --- centrist business weekly \\
Le HuffPost & $-$0.25 & 2012, French HuffPost --- progressive & CNews & 0.0 & i-Télé 1999 --- centrist; shift via D1 \\
Le Monde & $-$0.25 & 1944, Beuve-Méry --- centre-left & La Croix & $+$0.25 & 1880, Assomptionists --- Catholic \\
Le Nouvel Obs & $-$0.25 & 1964, Daniel --- non-communist left & Les Echos & $+$0.25 & 1908 --- liberal business daily \\
Marianne & $-$0.25 & 1997, J.-F.\ Kahn --- left-republican & Le Point & $+$0.25 & 1972, Imbert --- Gaullist right \\
L'Express & $-$0.25 & 1953, Servan-Schreiber --- centre-left & JDD & $+$0.25 & 1948, Lazareff --- centre-right; pre-Bolloré \\
Slate.fr & 0.0 & 2009, French offshoot of US Slate & Atlantico & $+$0.25 & 2011, Ferjou --- liberal right \\
Franceinfo & 0.0 & FTV/Radio France --- neutral remit & L'Opinion & $+$0.25 & 2013, Beytout --- pro-business \\
BFMTV & 0.0 & 2005 --- commercial 24h news & Le Figaro & $+$0.5 & 1826 --- conservative \\
TF1 INFO & 0.0 & TF1 digital; privatised 1987 & Causeur & $+$0.5 & 2007, É.\ Lévy --- conservative-right \\
20 Minutes & 0.0 & 2002 --- free mass-distribution daily & Valeurs actuelles & $+$0.5 & 1966 --- far-right \\
Le Parisien & 0.0 & 1944 --- popular Paris daily & Fdesouche & $+$0.5 & 2006 --- far-right aggregator \\
\bottomrule
\end{tabularx}
\end{nolinenumbers}
\end{table*}

\section{D7 Lexical Scoring: Full Results and Term Selection}
\label{appendix:d7}

This appendix reports the D7 family in full: per-outlet scores (\ref{appendix:d7scores}), lexicon construction and rejection log (\ref{appendix:d7reject}), and the embedding-probe stress test (\ref{appendix:d7probe}).

\subsection{Outlet-Level D7 Scores}
\label{appendix:d7scores}

Table~\ref{tab:d7full} gives D7\textsubscript{raw} for all 30 outlets on the all-category 2024--2025 corpus, with the underlying left- and right-signal article shares and per-outlet scored-article counts. Every outlet clears the 10-signal-article threshold on this window (30/30 coverage).

\begin{table*}[t]
\begin{nolinenumbers}
\centering
\small
\setlength{\tabcolsep}{3pt}
\caption{D7 lexical scores, all 30 outlets (2024--2025 corpus), two panels. D7\textsubscript{raw} = ($n_{\text{right}} - n_{\text{left}}$)/($n_{\text{right}} + n_{\text{left}}$); L\% and R\% are shares of articles whose plain-use occurrences for that side outnumber qualified-use ones. n\_art.\ counts scored articles; articles with unusable body text are dropped, so the column sums just below the 568{,}906 sampled total.}
\label{tab:d7full}
\begin{tabularx}{\textwidth}{@{}>{\RaggedRight\arraybackslash}Xrrrr | >{\RaggedRight\arraybackslash}Xrrrr@{}}
\toprule
Outlet & n\_art. & L\% & R\% & D7\_raw & Outlet & n\_art. & L\% & R\% & D7\_raw \\
\midrule
L'Humanité & 24,367 & 6.3\% & 0.6\% & $-$0.829 & Challenges & 16,980 & 1.5\% & 1.1\% & $-$0.145 \\
Mediapart & 9,234 & 1.0\% & 0.1\% & $-$0.784 & Le Parisien & 25,000 & 0.6\% & 0.5\% & $-$0.137 \\
Libération & 25,000 & 1.9\% & 0.5\% & $-$0.547 & TF1 INFO & 25,000 & 1.1\% & 0.9\% & $-$0.110 \\
Blast & 959 & 12.5\% & 3.8\% & $-$0.538 & Le Point & 24,224 & 1.3\% & 1.1\% & $-$0.098 \\
Le HuffPost & 19,841 & 1.8\% & 0.6\% & $-$0.535 & RFI & 24,358 & 1.3\% & 1.1\% & $-$0.078 \\
Slate.fr & 5,543 & 2.6\% & 0.8\% & $-$0.518 & L'Express & 18,805 & 2.3\% & 2.3\% & $-$0.005 \\
Le Nouvel Obs & 24,004 & 1.8\% & 0.6\% & $-$0.471 & France 24 & 10,408 & 2.4\% & 2.4\% & +0.002 \\
20 Minutes & 25,000 & 0.9\% & 0.4\% & $-$0.357 & L'Opinion & 14,913 & 0.8\% & 0.9\% & +0.079 \\
Ouest-France & 25,000 & 0.3\% & 0.1\% & $-$0.346 & Le Figaro & 25,000 & 0.5\% & 0.6\% & +0.144 \\
Marianne & 12,196 & 2.2\% & 1.1\% & $-$0.320 & CNews & 25,000 & 0.6\% & 0.9\% & +0.147 \\
La Croix & 25,000 & 1.6\% & 0.8\% & $-$0.316 & Atlantico & 6,411 & 3.7\% & 5.6\% & +0.195 \\
Franceinfo & 25,000 & 1.6\% & 0.9\% & $-$0.290 & Causeur & 4,747 & 5.0\% & 11.1\% & +0.375 \\
Le Monde & 25,000 & 1.1\% & 0.6\% & $-$0.274 & Valeurs actuelles & 19,258 & 1.3\% & 4.5\% & +0.540 \\
BFMTV & 25,000 & 1.2\% & 0.7\% & $-$0.246 & JDD & 22,036 & 1.5\% & 5.1\% & +0.546 \\
Les Echos & 25,000 & 1.1\% & 0.8\% & $-$0.169 & Fdesouche & 10,613 & 0.8\% & 2.6\% & +0.554 \\
\bottomrule
\end{tabularx}
\end{nolinenumbers}
\end{table*}

\subsection{Term Selection Log}
\label{appendix:d7reject}

From roughly 80 candidates, terms were culled manually against three operationalizable criteria; no automated correlation search was run and no terms were added post-hoc. \emph{Neutral saturation}: fires at high, roughly equal rates spectrum-wide, journalistically unmarked. \emph{Polysemy}: domain-specific use (religious, legal) fires reliably in ideologically opposed outlets. \emph{Wire contamination}: enters opposing outlets via AFP/official-language import, not editorial endorsement. Retained terms were validated by endorsed firing rates across five left- and six right-cluster outlets on the 2024--2025 corpus (median discrimination 4.6$\times$; full per-term ratios in the replication package). Raw occurrence rates are not used: they conflate plain and distanced use.

\paragraph{Rejected terms.} Grouped by rejection criterion, with the outlet at which the term fired at high rate despite an opposing editorial stance:

\begin{itemize}[leftmargin=1.2em,itemsep=1pt,topsep=2pt,parsep=0pt]
\item \emph{Event saturation.} \emph{émeutiers} (`rioters')---coverage-heavy outlets during the May 2024 Nouvelle-Calédonie unrest; lexically neutral.
\item \emph{Neutral saturation.} \emph{délinquance} (`delinquency'), \emph{criminalité} (`criminality')---Le Monde, La Croix (factual social reporting); \emph{flux migratoires} (`migratory flows')---all outlets (demographic register); \emph{souveraineté nationale} (`national sovereignty'), \emph{identité nationale} (`national identity')---spectrum-wide (constitutional/EU discourse); \emph{sans-papiers} (`undocumented migrants'), \emph{évasion fiscale} (`tax evasion')---Les Echos, Ouest-France (business/civic reporting).
\item \emph{Polysemy.} \emph{islamisation} (`Islamization'), \emph{séparatisme} (`separatism'), \emph{communautarisme} (`communitarianism')---La Croix (religious-affairs journalism); \emph{assistan[ct]}---all outlets (\emph{assistant social} `social worker' collocation).
\item \emph{Wire contamination.} \emph{clandestins} (`illegal migrants')---La Croix (AFP wire / official language).
\end{itemize}

\subsection{Signal Validation: Embedding Probe}
\label{appendix:d7probe}

\begin{figure}[tbp]
\begin{nolinenumbers}
\centering
\includegraphics[width=\columnwidth]{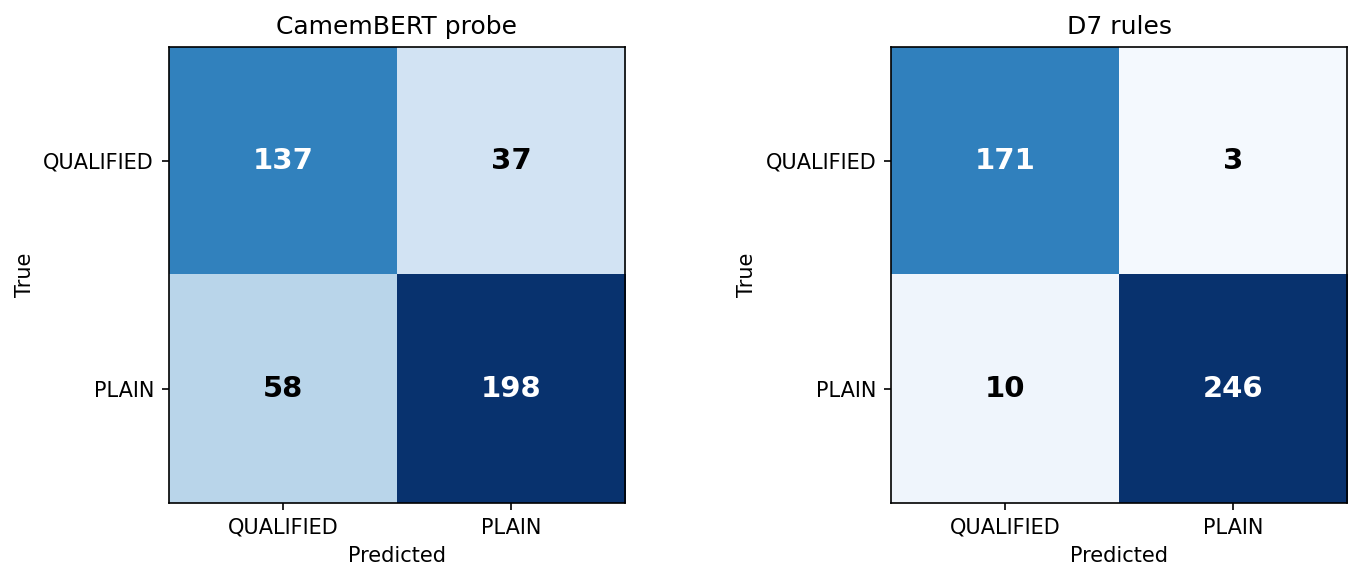}
\caption{Confusion matrices for the CamemBERT linear probe (left) and D7 rules (right) on the 430-occurrence PLAIN/QUALIFIED gold standard.}
\label{fig:d7probe}
\end{nolinenumbers}
\end{figure}

Figure~\ref{fig:d7probe} compares a logistic-regression probe on CamemBERT-base embeddings against D7 rules on the 430-occurrence gold standard (5-fold stratified CV). D7 reaches macro F1\,=\,0.969 with precision\,=\,0.988 for PLAIN. The probe reaches 0.775, and an ensemble underperforms D7 alone (0.856). The confusion matrices show why: D7's errors are predominantly conservative (10 false QUALIFIED against 3 false PLAIN), the probe's bidirectional. UMAP corroborates this --- no term, retained or rejected, shows a clean outlet-ideology axis (per-term $|\rho| \lesssim 0.4$). Correlations and UMAP plots: replication package.

\end{document}